\documentclass{article}

\usepackage{arxiv}
\usepackage{float} 
\usepackage[utf8]{inputenc} 
\usepackage[T1]{fontenc}    
\usepackage{hyperref}       
\usepackage{url}            
\usepackage{booktabs}       
\usepackage{amsfonts}       
\usepackage{nicefrac}       
\usepackage{microtype}      
\usepackage{lipsum}		
\usepackage{graphicx}
\usepackage{natbib}
\usepackage{doi}
\usepackage{tabularx} 
\usepackage{booktabs} 
\usepackage{multirow}
\usepackage{algpseudocode}
\usepackage{amsmath}
\usepackage{adjustbox}
\usepackage[ruled,vlined]{algorithm2e}

\title{MambaOutRS: A Hybrid CNN-Fourier Architecture for Remote Sensing Image Classification}

\author{
    Minjong Cheon \\
    KAIST Applied Science Research Institute \\
    Daejeon, South Korea \\
    \texttt{jmj2316@kaist.ac.kr} \\
    \And
    Changbae Mun\thanks{Corresponding author} \\
    Department of Electrical, Electronic \& Communication Engineering \\
    Hanyang Cyber University \\
    Seoul 04764, Republic of Korea \\
    \texttt{changbae@hycu.ac.kr}
}

\renewcommand{\shorttitle}{\textit{arXiv} Template}

\hypersetup{
pdftitle={A template for the arxiv style},
pdfsubject={q-bio.NC, q-bio.QM},
pdfauthor={Minjong},
pdfkeywords={First keyword, Second keyword, More},
}

\begin{document}
\maketitle

\begin{abstract}
Recent advances in deep learning for vision tasks have seen the rise of State Space Models (SSMs) like Mamba, celebrated for their linear scalability. However, their adaptation to 2D visual data often necessitates complex modifications that may diminish efficiency. In this paper, we introduce MambaOutRS, a novel hybrid convolutional architecture for remote sensing image classification that re-evaluates the necessity of recurrent SSMs. MambaOutRS builds upon stacked Gated CNN blocks for local feature extraction and introduces a novel Fourier Filter Gate (FFG) module that operates in the frequency domain to capture global contextual information efficiently. Our architecture employs a four-stage hierarchical design and was extensively evaluated on challenging remote sensing datasets: UC Merced, AID, NWPU-RESISC45, and EuroSAT. MambaOutRS consistently achieved state-of-the-art (SOTA) performance across these benchmarks. Notably, our MambaOutRS-t variant (24.0M parameters) attained the highest F1-scores of 98.41\% on UC Merced and 95.99\% on AID, significantly outperforming existing baselines, including larger transformer models and Mamba-based architectures, despite using considerably fewer parameters. An ablation study conclusively demonstrates the critical role of the Fourier Filter Gate in enhancing the model's ability to capture global spatial patterns, leading to robust and accurate classification. These results strongly suggest that the complexities of recurrent SSMs can be effectively superseded by a judicious combination of gated convolutions for spatial mixing and frequency-based gates for spectral global context. Thus, MambaOutRS provides a compelling and efficient paradigm for developing high-performance deep learning models in remote sensing and other vision domains, particularly where computational efficiency is paramount.

\end{abstract}

\section{Introduction}\label{sec:intro}
Remote sensing image classification (RSIC) stands as a significant and foundational task within the broader domain of Earth observation. Its successful implementation enables a diverse array of critical applications, including precise land use and land cover (LULC) mapping, comprehensive environmental monitoring, strategic urban planning, rapid disaster assessment, and efficient agricultural management \cite{wang2022machine}. 

Historically, RSIC relied predominantly on traditional machine learning methodologies. However, the exponential increase in remotely sensed data, stemming from a proliferation of diverse onboard sensors, has rendered manual interpretation and classification processes increasingly inefficient and impractical \cite{song2025pure}. The advent of deep learning (DL) paradigms, particularly Convolutional Neural Networks (CNNs) and, more recently, Transformer architectures, has catalyzed significant advancements in classification accuracy \cite{cheon2024combining}\cite{chen2024rsmamba}\cite{gupta2021remote}\cite{lv2022review}\cite{cheon2024kolmogorov}.  Especially, transformers have become foundational architectures across a myriad of visual tasks, including object detection, semantic segmentation, and video understanding \cite{qiu2025vivit}. However, their core attention mechanism, which scales quadratically with sequence length, presents significant scalability challenges and leads to computationally intensive inference, particularly when dealing with high-resolution images or extended sequences.   

In response to these limitations, the Mamba architecture has emerged as a promising alternative, leveraging State Space Models (SSMs) to achieve near-linear scalability in sequence length and facilitate hardware-efficient processing \cite{gu2023mamba}. Mamba's inherent efficiency, stemming from its 1D causal sequence modeling, offers linear scalability over Transformers. However, adapting it for non-causal 2D visual data necessitates significant modifications like bidirectional scanning and positional embeddings. This architectural overhead can diminish Mamba's efficiency or require substantial task-specific adjustments, suggesting simpler architectures might be more efficient for certain visual tasks. This challenge has consequently driven the development of specialized architectures such as Vision Mamba and MambaVision \cite{liu2024vision}\cite{hatamizadeh2025mambavision}.

Especially for remote sensing, RSMamba introduces a "dynamic multi-path activation mechanism" to overcome the inherent limitation of the vanilla Mamba, and RSMamba claims superior performance across multiple remote sensing image classification datasets \cite{chen2024rsmamba}. Despite the reported successes of models like RSMamba, a critical counter-argument emerges from the "MambaOut" model \cite{yu2025mambaout}. MambaOut conceptually posits that the Mamba architecture is ideally suited for tasks characterized by "long-sequence" and "autoregressive" (or causal token mixing) properties.

To summarize, the main contributions of this work are:
\begin{itemize}
\item We propose MambaOutRS, a novel hybrid convolutional architecture that effectively combines local spatial feature extraction via Gated CNN blocks with global frequency-domain filtering. This design challenges the necessity of recurrent State Space Models (SSMs) for remote sensing image classification.
\item We introduce the Fourier Filter Gate (FFG), a novel module that operates in the frequency domain to learn a frequency-wise gating mask, enhancing the model's ability to capture and process global contextual information efficiently.
\item We demonstrate that MambaOutRS achieves state-of-the-art performance on multiple remote sensing datasets (UC Merced, AID, EuroSAT, and NWPU-RESISC45), significantly outperforming larger transformer and Mamba-based models, while maintaining a competitive or even smaller parameter footprint.
\end{itemize}

\section{Method}\label{sec:method}

\subsection{MambaOut Architecture}

MambaOut is a convolutional vision backbone that removes the recurrent \textit{State Space Model} (SSM) used in Mamba~\cite{gu2023mamba}, and instead builds its architecture purely from stacked Gated CNN blocks~\cite{dauphin2017language}. It is designed to test the hypothesis that SSM is unnecessary for image classification tasks such as ImageNet.

\subsubsection{Gated CNN Block}

Each Gated CNN block follows a residual-style structure and consists of a dual linear path combined with a depthwise convolution over a portion of channels. Given an input tensor \( \mathbf{X} \in \mathbb{R}^{B \times H \times W \times C} \), the operations are as follows:

\begin{algorithm}[H]
\caption{Gated CNN Block}
\label{alg:gated_cnn}
\begin{algorithmic}[1]
\Require Input tensor $\mathbf{X} \in \mathbb{R}^{B \times H \times W \times C}$
\Ensure Output tensor $\mathbf{X}_{\text{out}} \in \mathbb{R}^{B \times H \times W \times C}$

\State $\mathbf{X}' \gets \texttt{Norm}(\mathbf{X})$ \Comment{Normalization}

\State $\mathbf{U} \gets \texttt{Linear}_1(\mathbf{X}') \in \mathbb{R}^{B \times H \times W \times 2h}$ \Comment{First projection}

\State $[\mathbf{g}, \mathbf{i}, \mathbf{c}] \gets \texttt{Split}(\mathbf{U})$ \Comment{Split into gate, identity, conv path}

\State $\mathbf{c}' \gets \texttt{DWConv}(\mathbf{c})$ \Comment{Depthwise conv on partial channel}

\State $\mathbf{Z} \gets \sigma(\mathbf{g}) \odot \texttt{Concat}(\mathbf{i}, \mathbf{c}')$ \Comment{Gated fusion}

\State $\mathbf{Y} \gets \texttt{Linear}_2(\texttt{GELU}(\mathbf{Z}))$ \Comment{Second projection}

\State $\mathbf{X}_{\text{out}} \gets \mathbf{X} + \mathbf{Y}$ \Comment{Residual connection}
\end{algorithmic}
\end{algorithm}

In abstract form, this block can be described as:
\[
\mathbf{X}_{\text{out}} = \mathbf{X} + \left( \text{TokenMixer}(\mathbf{X}' \mathbf{W}_1) \odot \sigma(\mathbf{X}' \mathbf{W}_2) \right) \mathbf{W}_3
\]
where the \texttt{TokenMixer} for MambaOut is a depthwise convolution:
\[
\text{TokenMixer}_{\text{GatedCNN}}(\mathbf{Z}) = \text{DWConv}(\mathbf{Z})
\]
and in contrast, Mamba uses:
\[
\text{TokenMixer}_{\text{Mamba}}(\mathbf{Z}) = \text{SSM}(\sigma(\text{Conv}(\mathbf{Z})))
\]

\subsubsection{Hierarchical Backbone Structure}

MambaOut employs a four-stage hierarchical design similar to ResNet and ConvNeXt. The full architecture is defined as:

\begin{center}
\texttt{Stem} \(\rightarrow\) [Stage 1: Gated CNN \(\times N_1\)] \(\rightarrow\) Downsample \(\rightarrow\) [Stage 2: Gated CNN \(\times N_2\)] \(\rightarrow\) Downsample \(\rightarrow\) \ldots \(\rightarrow\) Global Avg Pooling \(\rightarrow\) MLP Head
\end{center}

Each stage increases the number of channels and decreases spatial resolution. Gated CNN blocks can be stacked with varied depth depending on the desired model size (e.g., Tiny, Small, Base).

This design demonstrates that convolutional token mixing, when properly structured, can match or exceed the performance of SSM-based models in vision tasks such as ImageNet classification.

\subsection{Fourier Filter Gate}

To selectively emphasize or suppress specific frequency components, we introduce a novel module called the FourierFilterGate. This module operates in the frequency domain and learns a frequency-wise gating mask that is applied to the real-valued 2D Fourier transform of the input tensor.

\begin{algorithm}[H]
\caption{FFG Module}
\label{alg:ffg}
\begin{algorithmic}[1]
\Require Input tensor $\mathbf{x} \in \mathbb{R}^{B \times H \times W \times C}$
\Ensure Output tensor $\mathbf{y} \in \mathbb{R}^{B \times H \times W \times C}$
\State $\mathbf{x}' \gets \texttt{permute}(\mathbf{x}) \in \mathbb{R}^{B \times C \times H \times W}$
\State $\hat{\mathbf{x}} \gets \texttt{rfft2}(\mathbf{x}', \texttt{norm='ortho'}) \in \mathbb{C}^{B \times C \times H \times (W/2 + 1)}$
\State $\mathbf{g} \gets \sigma(\mathbf{w}) \in [0,1]^{1 \times C \times H \times (W/2 + 1)}$ \Comment{Learnable gate}
\State $\hat{\mathbf{x}}_{\text{gated}} \gets \hat{\mathbf{x}} \odot \mathbf{g}$ \Comment{Element-wise multiplication}
\State $\mathbf{x}_{\text{filtered}} \gets \texttt{irfft2}(\hat{\mathbf{x}}_{\text{gated}}, s=(H,W), \texttt{norm='ortho'})$
\State $\mathbf{y} \gets \texttt{permute}(\mathbf{x}_{\text{filtered}}) \in \mathbb{R}^{B \times H \times W \times C}$
\end{algorithmic}
\end{algorithm}

\subsection{Fourier Gate Block}

We embed the \texttt{FourierFilterGate} within a residual-style structure, forming the Fourier Gate Block (FGB). This block follows the general architecture of Transformer-based encoders, composed of two normalization layers, a frequency filtering module, and a feed-forward MLP.

Given input \( \mathbf{x} \in \mathbb{R}^{B \times H \times W \times C} \), the forward computation proceeds as:

\begin{align*}
\mathbf{z}_1 &= \texttt{FourierFilterGate}(\texttt{Norm}_1(\mathbf{x})) \\
\mathbf{x}_1 &= \mathbf{x} + \texttt{DropPath}(\mathbf{z}_1) \\
\mathbf{z}_2 &= \texttt{MLP}(\texttt{Norm}_2(\mathbf{x}_1)) \\
\mathbf{y} &= \mathbf{x}_1 + \texttt{DropPath}(\mathbf{z}_2)
\end{align*}

\noindent
This design allows the network to capture global context in the frequency domain while leveraging residual connections and non-linear transformation for enhanced expressiveness. The entire block is fully differentiable and compatible with end-to-end training.

\subsection{MambaOutRS}

We propose MambaOutRS, a hybrid convolutional architecture that integrates both spatial and frequency-domain inductive biases through the combination of Gated CNN blocks and Fourier Gate Blocks. The design follows a hierarchical four-stage structure similar to MambaOut \cite{yu2025mambaout}, but replaces conventional residual blocks with our proposed frequency-aware and gated token mixing modules.

The model processes an input tensor \( \mathbf{x} \in \mathbb{R}^{B \times H \times W \times C} \) through the following stages:

\begin{algorithm}[H]
\caption{MambaOutRS Forward Pipeline}
\label{alg:pipeline}
\begin{algorithmic}[1]
\Require Input image tensor $\mathbf{x} \in \mathbb{R}^{B \times H \times W \times C}$
\Ensure Output prediction logits $\mathbf{y} \in \mathbb{R}^{B \times K}$

\State $\mathbf{x}_0 \gets \texttt{Stem}(\mathbf{x})$ \Comment{Initial projection via convolution}

\State $\mathbf{x}_1 \gets \texttt{GatedCNNBlocks}(\mathbf{x}_0)$ \Comment{Stage 1: Local feature extraction}

\State $\mathbf{x}_2 \gets \texttt{Downsample}(\mathbf{x}_1)$
\State $\mathbf{x}_2 \gets \texttt{FourierGateBlocks}(\mathbf{x}_2)$ \Comment{Stage 2: Frequency-aware filtering}

\State $\mathbf{x}_3 \gets \texttt{Downsample}(\mathbf{x}_2)$
\State $\mathbf{x}_3 \gets \texttt{FourierGateBlocks}(\mathbf{x}_3)$ \Comment{Stage 3: Global receptive field}

\State $\mathbf{x}_4 \gets \texttt{GatedCNNBlocks}(\mathbf{x}_3)$ \Comment{Stage 4: Feature refinement}

\State $\mathbf{x}_{\text{norm}} \gets \texttt{LayerNorm}(\mathbf{x}_4)$
\State $\mathbf{x}_{\text{gap}} \gets \texttt{GlobalAvgPool}(\mathbf{x}_{\text{norm}})$
\State $\mathbf{y} \gets \texttt{MLPHead}(\mathbf{x}_{\text{gap}})$ \Comment{Final classification}
\end{algorithmic}
\end{algorithm}

This design allows MambaOutRS to effectively blend local spatial information with global frequency representations, achieving high representational power while maintaining efficiency. By removing the recurrent SSM from Mamba-based models and replacing it with lightweight convolutional and spectral modules, our architecture could achieve both efficiency and accuracy. The table~\ref{tab:mambaoutrs_configs} details the architectural configurations of the MambaOutRS variants (femto, kobe, and tiny), specifying the number of Gated CNN blocks in each of its four stages (Depths) and the channel dimensions at each stage (Dims).

\begin{table}[H]
\centering
\caption{Configurations of MambaOutRS Variants.}
\label{tab:mambaoutrs_configs}
\begin{tabular}{lcc}
\toprule
\textbf{Variant} & \textbf{Depths (Number of Blocks per Stage)} & \textbf{Dims (Channels per Stage)} \\
\midrule
MambaOutRS-f (femto) & [3, 3, 9, 3] & [48, 96, 192, 288] \\
MambaOutRS-k (kobe) & [3, 3, 15, 3] & [48, 96, 192, 288] \\
MambaOutRS-t (tiny) & [3, 3, 9, 3] & [96, 192, 384, 576] \\
\bottomrule
\end{tabular}
\end{table}

\begin{figure}[htbp]
    \centering
    \includegraphics[width=\linewidth]{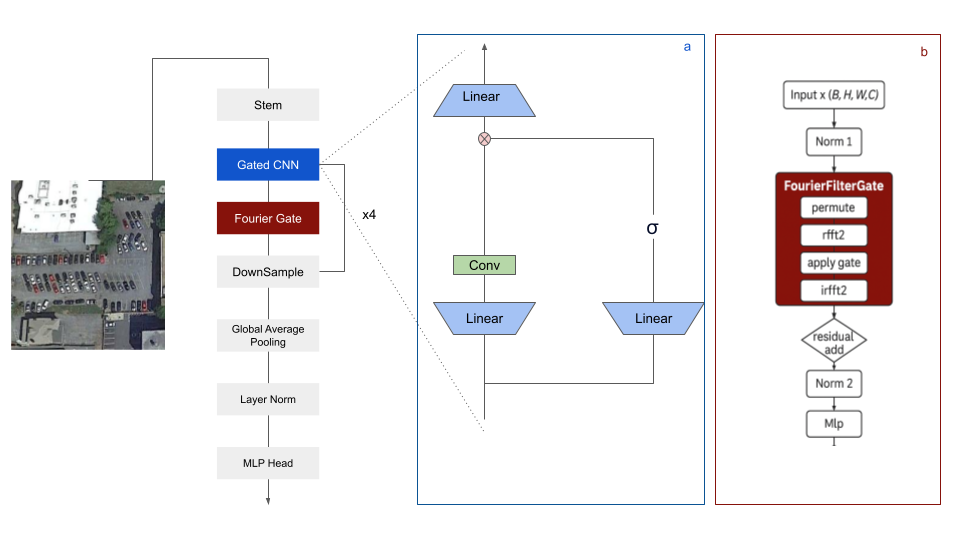} 
    \caption{
    Overview of the proposed architecture integrating Gated CNN and FourierFilterGate modules.
    Given an input image, the model first processes features through a stem layer followed by a sequence of Gated CNN and Fourier Gate modules repeated four times. 
    Panel \textbf{(a)} shows the internal structure of the Gated CNN block, which employs a gated fusion mechanism between two linear paths and a convolutional branch, where the gating signal \( \sigma \) modulates feature integration. 
    Panel \textbf{(b)} details the FourierFilterGate module: it performs a real-valued 2D FFT, applies a learnable frequency-domain gating mask, and reconstructs the filtered signal via inverse FFT, enhancing spectral selectivity. 
    Downstream components include downsampling, global average pooling, layer normalization, and an MLP head for final prediction.
    }
    \label{fig:architecture}
\end{figure}

\section{Experimental Results and Analysis}
\subsection{Dataset Description}

To ensure a fair comparison between our proposed model, MambaOutRS, and RSMamba, we evaluated both models on the same datasets used in the RSMamba experiments. Specifically, three benchmark datasets were selected: UC Merced, AID, and NWPU-RESISC45. In addition, we included EuroSAT, a widely used dataset for remote sensing image classification, to further validate the generalization performance.

\begin{itemize}
    \item \textbf{Aerial Image Dataset (AID)} \cite{xia2017aid}: A larger collection, AID features 30 classes, with 200 to 400 images per class, accumulating to between 6,000 and 12,000 high-resolution 600x600 RGB images. These images were derived from Google Earth imagery, though they are preprocessed and not freely available.
    
    \item \textbf{EuroSAT} \cite{helber2019eurosat}: Offering a multi-spectral perspective, EuroSAT comprises 27,000 labeled and geo-referenced images across 10 land use and land cover classes. These 64x64 pixel images are derived from Sentinel-2 satellite data and cover 13 spectral bands, including visible (RGB), near-infrared, and shortwave-infrared parts of the spectrum.
    
    \item \textbf{NWPU-RESISC45} \cite{cheng2017remote}: This dataset contains 31,500 images distributed across 45 classes, with 700 images per class. The images are 256x256 pixels in RGB and are also sourced from Google Earth.
    
    \item \textbf{UC Merced (UCM) Land Use Dataset} \cite{yang2010bag}: This dataset consists of 2,100 images, categorized into 21 land use and land cover classes, with 100 images per class. Each image is 256x256 pixels in RGB color space and was extracted from the USGS National Map Urban Area Imagery collection.
\end{itemize}

\begin{table}[htbp]
\centering
\caption{Summary of aerial and satellite image classification datasets used in our experiments.}
\begin{tabular}{lcccc}
\toprule
\textbf{Dataset} & \textbf{Source} & \textbf{Classes} & \textbf{Resolution} & \textbf{Channels} \\
\midrule
AID        & Google Earth         & 30  & 600$\times$600 & RGB \\
EuroSAT                      & Sentinel-2           & 10  & 64$\times$64   & 13-band MS / RGB \\
NWPU-RESISC45                & Google Earth         & 45  & 256$\times$256 & RGB \\
UC Merced (UCM)              & USGS National Map    & 21  & 256$\times$256 & RGB \\
\bottomrule
\end{tabular}
\label{tab:dataset_summary}
\end{table}

\subsection{Training Details}

We implement our model using PyTorch and train all variants for a total of 25 epochs. Evaluation is performed on the test set using the checkpoint that yields the best validation loss.

In our paper, we employ a fixed input image size of $224 \times 224$ and implement data augmentation techniques including random cropping, flipping, photometric distortion, Mixup, and CutMix to improve generalization. Optimization is performed using the Adam optimizer, and the learning rate is initialized to 0.001.

\subsection{Comparison with SOTA models}
Table~\ref{tab:performance_comparison} presents a comprehensive comparison of various backbone models across the UC Merced, AID, and RESISC45 datasets. Our proposed models, collectively denoted as \textit{MambaOutRS}, consistently achieve state-of-the-art (SOTA) performance on UC Merced and AID datasets. Notably, MambaOutRS-t attains the highest F1-scores of 98.41 and 95.99 on UC Merced and AID, respectively, outperforming all existing baselines including large-scale transformer and Mamba variants.

The naming convention of MambaOutRS variants reflects their relative scale: f (femto), k (kobe), and t (tiny), corresponding to 6.1M, 8.0M, and 24.0M parameters, respectively. Despite using significantly fewer parameters than RSMamba-H (33.1M), our MambaOutRS-t and even lightweight variants such as MambaOutRS-k outperform RSMamba across two datasets. MambaOutRS-f, the smallest variant, still maintains competitive performance, outperforming much larger models such as ResNet-101 and Swin-B on multiple benchmarks.

While RSMamba-H achieves the top score on RESISC45, MambaOutRS models deliver comparable accuracy with far fewer parameters. These results highlight the strength of our architecture in achieving high accuracy under constrained model budgets, demonstrating a favorable trade-off between model capacity and performance.

\begin{table*}[!h]
\centering
\caption{Performance comparison of different models on UC Merced, AID, and RESISC45 datasets. Metrics reported are Precision (P), Recall (R), and F1-score (F1). Bold values indicate the best performance. Results for ResNet, ViT, Swin, and RSMamba variants are taken from \cite{chen2024rsmamba}.}
\label{tab:performance_comparison}
\resizebox{\textwidth}{!}{%
\begin{tabular}{lccccccccccc}
\toprule
\multirow{2}{*}{\textbf{Method}} & \multirow{2}{*}{\textbf{Params (M)}} & \multicolumn{3}{c}{\textbf{UC Merced}} & \multicolumn{3}{c}{\textbf{AID}} & \multicolumn{3}{c}{\textbf{RESISC45}} \\
\cmidrule(lr){3-5} \cmidrule(lr){6-8} \cmidrule(lr){9-11}
& & P & R & F1 & P & R & F1 & P & R & F1 \\
\midrule
ResNet-18 [6] & 11.2 & 87.98 & 87.46 & 87.40 & 88.70 & 88.17 & 88.30 & 88.73 & 88.44 & 88.45 \\
ResNet-50 [6] & 23.6 & 91.99 & 91.74 & 91.65 & 89.44 & 88.66 & 88.87 & 92.67 & 92.47 & 92.47 \\
ResNet-101 [6] & 42.6 & 92.40 & 92.22 & 92.12 & 91.03 & 90.63 & 90.81 & 92.75 & 92.57 & 92.56 \\
\midrule
DeiT-T [16] & 5.5 & 86.92 & 86.66 & 86.53 & 85.23 & 84.52 & 84.52 & 87.66 & 86.78 & 86.79 \\
DeiT-S [16] & 21.7 & 88.95 & 88.41 & 88.41 & 85.88 & 85.19 & 85.34 & 88.21 & 87.47 & 87.43 \\
DeiT-B [16] & 85.8 & 89.14 & 88.73 & 88.70 & 87.32 & 86.07 & 86.07 & 89.04 & 88.62 & 88.65 \\
ViT-B [7] & 88.3 & 91.09 & 90.79 & 90.77 & 89.39 & 88.65 & 88.86 & 88.84 & 88.65 & 88.62 \\
ViT-L [7] & 303.0 & 91.98 & 91.32 & 91.26 & 90.19 & 88.86 & 89.17 & 91.22 & 91.08 & 91.04 \\
Swin-T [8] & 27.5 & 90.87 & 90.63 & 90.40 & 86.49 & 85.66 & 85.77 & 90.15 & 90.06 & 90.06 \\
Swin-S [8] & 48.9 & 91.08 & 90.95 & 90.82 & 87.50 & 86.80 & 86.89 & 92.05 & 91.88 & 91.84 \\
Swin-B [8] & 86.8 & 91.85 & 91.74 & 91.62 & 89.84 & 89.01 & 89.07 & 93.63 & 91.58 & 93.56 \\
\midrule
Vim-Ti$^\dagger$ [14] & 7.0 & 89.06 & 88.73 & 88.68 & 87.76 & 86.98 & 87.13 & 89.24 & 89.02 & 88.97 \\
VMamba-T [15] & 30.0 & 93.14 & 92.85 & 92.81 & 91.59 & 90.94 & 91.10 & 93.97 & 93.96 & 93.94 \\
\midrule
RSMamba-B & 6.4 & 94.14 & 93.97 & 93.88 & 92.02 & 91.53 & 91.66 & 94.87 & 94.87 & 94.84 \\
RSMamba-L & 16.2 & 95.03 & 94.76 & 94.74 & 92.31 & 91.75 & 91.90 & 95.03 & 95.05 & 95.02 \\
RSMamba-H & 33.1 & 95.47 & 95.23 & 95.25 & 92.97 & 92.51 & 92.63 & \textbf{95.22} & \textbf{95.19} & \textbf{95.18} \\
\midrule
MambaOutRS-f (ours) & 6.1 & 98.27 & 98.10 & 98.10 & 91.20 & 90.58 & 90.60 & 91.20 & 90.58 & 90.60 \\
MambaOutRS-k (ours) & 8.0 & 97.62 & 97.46 & 97.44 & 94.04 & 93.90 & 93.90 & 92.32 & 92.04 & 92.06 \\
MambaOutRS-t (ours) & 24.0 & \textbf{98.53} & \textbf{98.41} & \textbf{98.41} & \textbf{96.07} & \textbf{96.00} & \textbf{95.99} & 94.30 & 94.10 & 94.09 \\
\bottomrule
\end{tabular}%
}
\end{table*}

\subsection{Ablation Study: Effect of Fourier Gate Block on Performance}

In this subsection, we investigate the impact of the FGB on the classification performance of our MambaOutRS variants across four benchmark datasets: UC Merced, AID, RESISC45, and EuroSAT. Table~\ref{tab:FGB_ablation} compares the results obtained with and without the inclusion of the FGB.

The results clearly demonstrate that integrating the FGB substantially improves model performance, particularly on the UC Merced, AID, and EuroSAT datasets. For instance, the largest variant, MambaOutRS-t (tiny), achieves an F1-score increase from 94.58\% without FGB to 98.41\% with FGB on UC Merced, and from 93.19\% to 95.99\% on AID. On EuroSAT, both MambaOutRS-t and MambaOutRS-f (femto) reach 98.30\% F1-score with FGB, significantly outperforming their respective F1-scores without FGB (94.58\% and 97.44\%). Similar improvements are observed for the MambaOutRS-k (kobe) variant, where FGB contributes consistently to gains in precision, recall, and F1-score.

Although the improvements on the RESISC45 dataset are comparatively smaller, the addition of FGB still yields measurable performance gains across all variants. These findings underscore the importance of the frequency-domain filtering mechanism implemented by the FGB, which enhances the model’s ability to capture global spatial patterns. This leads to more robust and accurate aerial image classification, especially for datasets where global context is critical.

\begin{table*}[!h]
\centering
\caption{Comparison of MambaOutRS variants with and without FGB on UC Merced, AID, RESISC45, and EuroSAT datasets.}
\label{tab:FGB_ablation}
\resizebox{\textwidth}{!}{%
\begin{tabular}{lccccccccccccc}
\toprule
\multirow{2}{*}{\textbf{Method}} & \multirow{2}{*}{\textbf{FGB}} & \multicolumn{3}{c}{\textbf{UC Merced}} & \multicolumn{3}{c}{\textbf{AID}} & \multicolumn{3}{c}{\textbf{RESISC45}} & \multicolumn{3}{c}{\textbf{EuroSAT}} \\
\cmidrule(lr){3-5} \cmidrule(lr){6-8} \cmidrule(lr){9-11} \cmidrule(lr){12-14}
& & P & R & F1 & P & R & F1 & P & R & F1 & P & R & F1 \\
\midrule
MambaOutRS-f & Yes & \textbf{98.27} & \textbf{98.10} & \textbf{98.10} & 91.20 & 90.58 & 90.60 & \textbf{91.20} & \textbf{90.58} & \textbf{90.60} & \textbf{98.30} & \textbf{98.30} & \textbf{98.30} \\
MambaOutRS-f & No  & 97.63 & 97.46 & 97.44 & \textbf{91.82} & \textbf{91.50} & \textbf{91.47} & 89.11 & 88.40 & 88.33 & 97.63 & 97.46 & 97.44 \\
\midrule
MambaOutRS-k & Yes & \textbf{97.62} & \textbf{97.46} & \textbf{97.44} & \textbf{94.04} & \textbf{93.90} & \textbf{93.90} & \textbf{92.32} & \textbf{92.04} & \textbf{92.06} & \textbf{97.63} & \textbf{97.60} & \textbf{97.61} \\
MambaOutRS-k & No  & 95.92 & 95.98 & 95.54 & 93.27 & 92.80 & 92.83 & 91.98 & 91.30 & 91.41 & 95.92 & 95.56 & 95.54 \\
\midrule
MambaOutRS-t & Yes & \textbf{98.53} & \textbf{98.41} & \textbf{98.41} & \textbf{96.07} & \textbf{96.00} & \textbf{95.99} & \textbf{94.30} & \textbf{94.10} & \textbf{94.09} & \textbf{98.30} & \textbf{98.30} & \textbf{98.30} \\
MambaOutRS-t & No  & 95.00 & 95.00 & 94.58 & 93.45 & 93.20 & 93.19 & 91.60 & 91.00 & 91.04 & 95.00 & 94.60 & 94.58 \\
\bottomrule
\end{tabular}%
}
\end{table*}

\section{Conclusion}
In this paper, we introduced MambaOutRS, a novel hybrid convolutional architecture designed for remote sensing image classification that eschews the recurrent SSM of Mamba in favor of a synergistic combination of Gated CNN blocks and a proposed FGB. Our primary hypothesis was that effective image classification could be achieved without SSMs by leveraging the strengths of local convolutional processing and global frequency-domain filtering.

Our experimental results robustly support this hypothesis. MambaOutRS variants consistently achieved state-of-the-art performance on challenging remote sensing datasets, including UC Merced, AID, and NWPU-RESISC45. Notably, MambaOutRS-t, with 24.0M parameters, set new benchmarks with F1-scores of 98.41\% on UC Merced and 95.99\% on AID. These results significantly surpass existing baselines, including larger transformer models like ViT-L (303.0M parameters) and Swin-B (86.8M parameters), as well as Mamba-based architectures such as VMamba-T and RSMamba-H, despite often utilizing a considerably lower parameter count. For instance, MambaOutRS-t outperformed RSMamba-H (33.1M parameters) on UC Merced and AID, demonstrating a superior trade-off between model capacity and performance. Even our smallest variant, MambaOutRS-f (6.1M parameters), exhibited competitive performance, outperforming models like ResNet-101.

The ablation study on the Fourier Gate Block provided crucial insights into its contribution. The integration of FGB consistently led to substantial performance improvements across all MambaOutRS variants and datasets, particularly on UC Merced and AID. For MambaOutRS-t, the FGB improved the F1-score from 94.58\% to 98.41\% on UC Merced and from 93.19\% to 95.99\% on AID. This validates our design choice to incorporate frequency-domain filtering, highlighting its efficacy in enabling the model to capture global contextual information that complements the local feature extraction capabilities of Gated CNN blocks. The FGB's ability to selectively emphasize or suppress specific frequency components proves vital for robust and accurate aerial image classification.

In conclusion, MambaOutRS establishes a new paradigm for efficient and high-performing vision backbones by demonstrating that the complexities of recurrent SSMs can be effectively replaced by a judicious combination of gated convolutions for spatial mixing and Fourier-based gates for spectral global context. This work opens promising avenues for developing lightweight yet powerful deep learning models for diverse vision tasks, especially in domains where computational resources are a constraint. Future work will explore the application of MambaOutRS to other remote sensing tasks, such as object detection and segmentation, and investigate further optimizations for real-time inference.

\clearpage
\bibliographystyle{unsrtnat}
\bibliography{template}
\end{document}